\documentclass[letterpaper, 10 pt, conference]{ieeeconf}

\IEEEoverridecommandlockouts
\usepackage{iftex}
\ifXeTeX
  \usepackage{fontspec}
\fi

\usepackage{amsmath}
\usepackage{amssymb}
\usepackage{booktabs}
\usepackage{multirow}
\usepackage{graphicx}
\usepackage{subcaption}
\usepackage{xcolor}
\usepackage{makecell}
\usepackage{placeins}
\usepackage{cite}

\makeatletter
\let\NAT@parse\relax
\makeatother
\usepackage[colorlinks=true,citecolor=red,linkcolor=red,urlcolor=red]{hyperref}

\title{\LARGE \bf   GeoFlow-SLAM++: A Robust Multi-Camera Visual-Inertial SLAM System with Relocalization}

\author{
Wei Feng$^{*}$, Tingyang Xiao$^{*}$, Liu Liu$^{\dagger}$, Xiaolin Zhou,
Zhizhong Su\\
\\
Horizon Robotics \quad
\\
$^{*}$Equal contribution \quad $^{\dagger}$Project Lead
}

\begin{document}

\maketitle
\thispagestyle{empty}
\pagestyle{empty}

\begin{abstract}
Monocular and RGB-D visual-inertial SLAM systems remain susceptible to limited field of view, sensor-specific failure modes, and unreliable cross-session relocalization. To address these issues, we present GeoFlow-SLAM++, a tightly coupled multi-camera visual-inertial SLAM system that extends GeoFlow-SLAM from a single RGB-D sensor to a calibrated multi-camera rig with a unified body-centric formulation. Within this multi-camera framework, GeoFlow-SLAM++ supports two interchangeable visual front-ends: a conventional ORB front-end and a neural network feature (NN-Feature) front-end built on SuperPoint and LightGlue. The system unifies tracking, mapping, and relocalization on a shared body state, and combines multi-camera reprojection constraints, IMU pre-integration, cross-view place recognition, and dual-stream optical flow/NN-Feature tracking for robust localization. As an optional extension, the system can further incorporate cross-view-consistent pseudo-depth predictions from RGB images as auxiliary geometric constraints. We evaluate GeoFlow-SLAM++ on EuRoC, OpenLORIS, TUM, Hilti, and a self-collected handheld multi-camera dataset. Results show that the NN-Feature front-end improves robustness in appearance-challenging scenarios, the multi-camera formulation achieves competitive localization accuracy on Hilti, and the unified cross-view relocalization design reaches LiDAR-comparable performance on the handheld dataset.
\end{abstract}

\section{Introduction}
\label{sec:introduction}
Simultaneous localization and mapping (SLAM) underpins autonomous navigation and accurate pose estimation in embodied robotics. Visual-inertial (VI) SLAM systems such as ORB-SLAM3~\cite{orbslam3}, S-VIO~\cite{svio}, VINS~\cite{vinsmono,vinsfusion,vinsrgbd}, and OKVIS~\cite{okvis} provide compact and mature solutions. Our previous GeoFlow-SLAM~\cite{geoflow_v1} improved RGB-D VI-SLAM accuracy and robustness by combining dual-stream optical flow feature tracking with generalized iterative closest point (GICP)~\cite{gicp} geometric constraints. However, a single RGB-D sensor still suffers from limited visual coverage and limited appearance evidence. In practice, active depth sensing can fail outdoors or on reflective surfaces, a single narrow field of view can easily lose trackable features under occlusion or degenerate motion, and relocalization remains fragile under large temporal or viewpoint changes.

To improve robustness beyond a single RGB-D sensor, GeoFlow-SLAM can be extended in two aspects. First, a calibrated multi-camera rig provides broader visual coverage and stronger redundancy for tracking, loop closure, and relocalization. Second, a dual visual front-end design supports both an ORB front-end and an NN-Feature front-end. The ORB front-end provides an efficient and standard solution, while the NN-Feature front-end targets stronger appearance matching under illumination, texture, and viewpoint changes.

In this paper, we present \textbf{GeoFlow-SLAM++}, a tightly coupled multi-camera VI-SLAM framework centered on a dual visual front-end design. Our main contributions are as follows:
  \begin{itemize}
      \item A \textbf{unified multi-camera SLAM framework} that extends GeoFlow-SLAM from a single RGB-D sensor to a calibrated multi-camera rig, where tracking, mapping, loop closure, and relocalization are all formulated on a shared body state.
      \item A \textbf{dual visual front-end design} that supports both a conventional ORB front-end and a learning-based NN-Feature front-end based on SuperPoint and LightGlue under the same downstream geometric interface.
      \item A \textbf{multi-camera tightly coupled optimization pipeline} that combines dual-stream visual tracking, multi-camera reprojection residuals, IMU pre-integration, and cross-view place recognition for robust localization and relocalization.
      \item An \textbf{optional pseudo-depth extension} that introduces cross-view-consistent pseudo-depth predictions from RGB images as supplementary geometric constraints in degraded scenarios.
  \end{itemize}
\section{Related Work}
\label{sec:related_work}
\subsection{Monocular VI-SLAM}
Monocular VI-SLAM integrates a single camera and an IMU via tightly coupled sliding-window optimization. Representative methods such as OKVIS~\cite{okvis} leverage keyframe-based formulations to jointly optimize visual and inertial residuals. Following this optimization paradigm, VINS-Mono~\cite{vinsmono} provides a canonical monocular implementation, whereas VINS-Fusion~\cite{vinsfusion} and ORB-SLAM3~\cite{orbslam3} extend the same optimization backbone to support diverse sensor setups and multi-map operations. Despite delivering mature and robust tightly coupled baselines, existing methods are inherently limited to a single viewing perspective. This leads to fundamental single-view vulnerabilities: restricted field of view, degraded observability under viewpoint-degenerate motions, and catastrophic tracking failures when the camera view is occluded or geometrically degraded.

\subsection{Multi-Camera VI-SLAM}
Multi-camera configurations expand the aggregate field of view and
significantly reduce the probability of simultaneous tracking
failures. To this end, OpenMAVIS~\cite{mavis} investigates tightly coupled VI-SLAM
tailored for partially overlapping multi-camera setups, whereas
Multicam-SLAM~\cite{multicamslam} addresses non-overlapping configurations for
visual localization and navigation. Though these prior works fully leverage the benefits of multi-camera redundancy, they lack a unified modeling strategy to integrate diverse multi-camera measurements into a cohesive optimization pipeline, which limits the overall accuracy and robustness of multi-camera fusion.

\subsection{LiDAR-Inertial and LiDAR-Visual-Inertial Fusion}
Range-based SLAM systems leverage direct distance measurements to establish
definitive geometric constraints. For example, FAST-LIO2~\cite{fastlio2}
provides a representative formulation for LiDAR-inertial odometry.
Moving a step further, R3LIVE~\cite{r3live}, FAST-LIVO2~\cite{fastlivo2}, and
OmniLIVO~\cite{omnilivo} combine LiDAR, visual, and inertial cues to achieve
dense LiDAR-visual-inertial odometry (LIVO) and scene reconstruction. Despite the superior robustness of existing LIVO systems in large-scale environments, they rely on heavy, power-consuming, and expensive hardware
payloads. Such inherent hardware dependence restricts their applicability in lightweight, low-power, and cost-sensitive scenarios.

\subsection{Neural Network Feature and Place Recognition}
The NN-Feature paradigm has remarkably improved local data association and global place
retrieval under drastic illumination, texture, and viewpoint alterations.
SuperPoint~\cite{superpoint} extracts repeatable keypoints alongside
local descriptors, while LightGlue~\cite{lightglue} delivers highly efficient
feature matching via deep neural networks. For global visual place
recognition, representative frameworks like NetVLAD~\cite{netvlad}, MixVPR~\cite{mixvpr}, and
BoQ~\cite{boq} generate compact image-level global descriptors, whereas
DBoW2~\cite{dbow2} remains the standard bag-of-words baseline for ORB-based
SLAM. Recent NN-Feature SLAM adaptations, such as SuperVINS~\cite{supervins},
have verified that deep local features effectively improve tracking robustness in challenging scenarios. Nevertheless, the practical applicability of these approaches is heavily constrained by sensor hardware configurations, stringent real-time requirements, and challenges in seamless integration with back-end optimization modules.

\subsection{3D Foundation Models for Geometry}
Feed-forward 3D foundation models, including the Depth Anything series~\cite{depthanything,depthanythingv2,depthanythingv3},
DUSt3R~\cite{dust3r}, MASt3R~\cite{mast3r}, and VGGT~\cite{vggt}, infer dense
depth, pixel-wise correspondence, or explicit 3D structures directly from
standard RGB images, with recent applications extending to downstream SLAM systems such as IRIS-SLAM~\cite{irisslam}. These foundation models offer an unprecedented opportunity to inject passive geometric constraints into VI estimation frameworks. Nevertheless, because these feed-forward networks predict geometry on individual frames or pairs, their outputs cannot guarantee the strict temporal and probabilistic consistency required by an online SLAM back-end. Therefore, we treat this learned geometry as a source of supplementary, consistency-checked constraints rather than a primary source of metric depth.

\begin{figure*}[!tbp]
\centering
\includegraphics[width=0.95\textwidth]{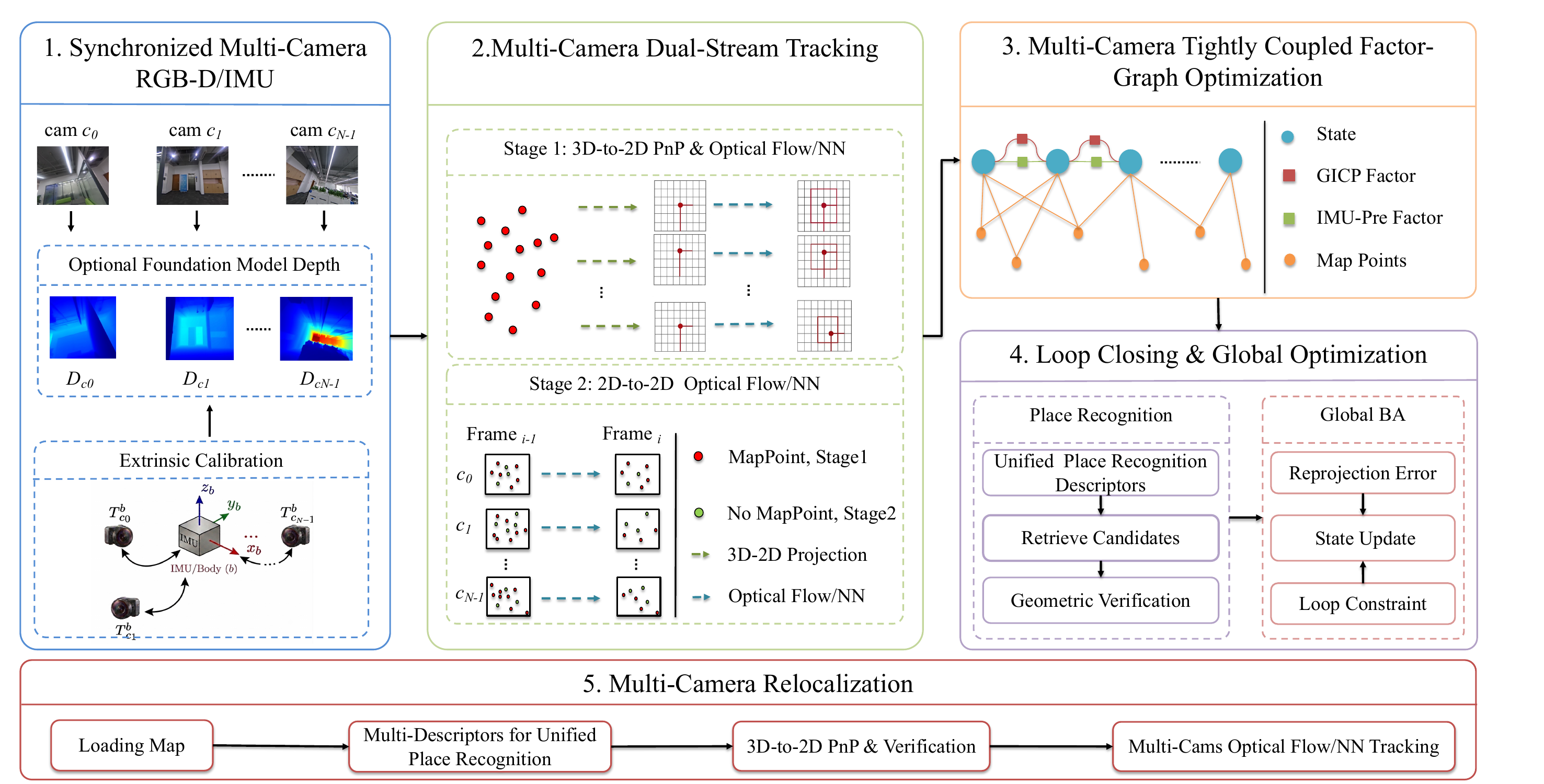}
\caption{\textbf{System architecture of GeoFlow-SLAM++.} The pipeline ingests synchronized image streams from a calibrated multi-camera rig alongside high-frequency IMU measurements. The front-end coordinates an ORB pipeline, an NN-Feature pipeline, and dual-stream optical flow tracking to establish robust data association through cross-view geometric verification. The back-end then solves a centralized factor graph that jointly optimizes the robot body state by fusing multi-camera reprojection residuals, IMU pre-integration, and depth-conditioned GICP constraints. The loop closure module aggregates individual per-camera descriptors $\mathbf{v}_{c_k}$ into unified global descriptors to trigger cross-view place recognition and 7-DoF relocalization against the global map.}

\label{fig:geoflow-slam++}
\end{figure*}

\section{System Overview}
\label{sec:system_overview}
GeoFlow-SLAM++ extends GeoFlow-SLAM from a single RGB-D sensor to a calibrated
multi-camera rig. As illustrated in the architectural diagram
Fig.~\ref{fig:geoflow-slam++}, the centralized pipeline ingests synchronized
image streams alongside high-frequency IMU measurements. Upon data arrival,
the system routes the images through one of two selected visual feature
front-ends:
\begin{itemize}
    \item The \textbf{ORB front-end} extracts ORB keypoints and generates bag-of-words descriptors for standard operations.
    \item The \textbf{NN-Feature front-end} extracts SuperPoint keypoints, establishes correspondences via LightGlue, and computes MixVPR descriptors for robust global visual place retrieval.
\end{itemize}

Both front-ends provide high-quality local features for frame-to-frame
tracking and sparse mapping, while simultaneously generating image-level
global descriptors to trigger loop closure and cross-session
relocalization.

To fully leverage multi-camera redundancy, the front-end treats the calibrated
multi-camera system as a unified source of tracking constraints. The centralized
body-state dual-stream optical flow/NN-Feature tracking aggregates valid data
associations from all camera viewpoints, minimizing the risk of tracking
failure caused by single-view occlusion or degenerate observations. The tracking module
then applies strict cross-view epipolar and triangulation checks to eliminate
geometrically inconsistent associations. When the optional depth module is
enabled, a multi-camera 3D foundation model predicts depth directly from the synchronized RGB images based on fixed extrinsics. Only geometrically consistent depth predictions are passed to downstream components
to prevent noisy artifacts from corrupting the state estimate.

The back-end relies on a centralized factor-graph optimizer. It jointly optimizes the robot body state by fusing multi-camera
reprojection residuals, IMU pre-integration terms, and GICP constraints
whenever RGB-D data or validated pseudo-depth predictions are
available. Loop closure and cross-session relocalization follow an identical
body-centric design. The loop closure module aggregates individual per-camera descriptors into a single, unified query vector, verifies
candidates through explicit cross-view geometric checks, and incorporates the
accepted loop constraints to update and optimize the global map.

\section{Method}
\label{sec:method}
\subsection{Dual Visual Feature Front-Ends}
\label{subsec:features}
GeoFlow-SLAM++ provides dual feature front-ends that share the same downstream
geometric modules. Both front-ends output keypoints, local descriptors, and
global descriptors; the ORB front-end prioritizes efficiency, while the
NN-Feature front-end targets stronger appearance matching under illumination,
texture, and viewpoint changes.

\noindent\textbf{ORB Front-End.} This efficiency-oriented front-end combines
hand-crafted FAST keypoints, rotated BRIEF descriptors, and bag-of-words place retrieval. It yields a local feature set
$\mathcal{F}_{c_k}^{\text{ORB}} = \{(\mathbf{p}_{c_k}^j,
\mathbf{d}_{c_k}^j)\}$. Loop closure retrieval is handled via a pre-trained
DBoW2 vocabulary, which emits per-camera bag-of-words vectors
$\mathbf{v}_{c_k}$ that are subsequently aggregated by the unified descriptor
operator $\mathcal{A}(\cdot)$ (detailed in Section~\ref{subsec:loop_closing}).

\noindent\textbf{NN-Feature Front-End.} This NN-Feature front-end fuses
three learning-based components: SuperPoint provides self-supervised,
illumination-invariant keypoints and local descriptors; LightGlue executes attention-driven matching with adaptive computation depth and supports
both local feature association and loop closure verification; MixVPR extracts a compact global descriptor $\mathbf{v}_{c_k}$ that is normalized and aggregated by $\mathcal{A}(\cdot)$ for visual place retrieval.

\noindent\textbf{Shared Feature Interface.} The dual front-ends hide their internal
feature implementations behind the same downstream interface. Each camera
provides a keypoint coordinate set $\{\mathbf{p}_{c_k}^j\}$, local descriptors
$\mathbf{d}_{c_k}^j$, and global descriptors $\mathbf{v}_{c_k}$.
Tracking and mapping consume the local features, while loop closure and
relocalization consume the global descriptors.

\begin{figure}[!tbp]
\centering
\begin{subfigure}{\columnwidth}
  \centering
  \includegraphics[width=\linewidth]{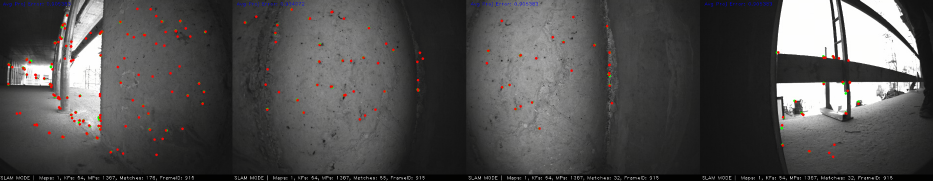}
  \caption{ORB front-end}
  \label{fig:feat_count_orb}
\end{subfigure}\\[2pt]
\begin{subfigure}{\columnwidth}
  \centering
  \includegraphics[width=\linewidth]{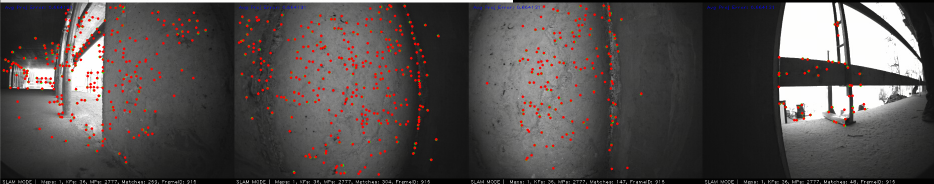}
  \caption{NN-Feature front-end}
  \label{fig:feat_count_nn}
\end{subfigure}
\caption{\textbf{Multi-camera feature extraction under the dual visual feature front-ends
on a representative multi-camera frame} (4-camera stitched view, identical
viewpoint, top-$500$ keypoint budget per camera). Red dots denote tracked
keypoints. (a) The ORB front-end tracks fewer features in textureless and
over-exposed regions. (b) The NN-Feature front-end tracks more in most views and
distributes them more evenly across low-textured areas.}
\label{fig:feat_count}
\end{figure}

Fig.~\ref{fig:feat_count} illustrates the difference between the two front-ends on a
representative multi-camera frame with the same budget of $500$ keypoints per
camera. The ORB front-end tracks fewer keypoints in low-texture or over-exposed
regions, while the NN-Feature front-end uses the budget more fully and spreads
features more evenly. 

\subsection{Dual-stream Optical Flow/NN-Feature Tracking}
\label{subsec:flow}
The tracker combines 3D-to-2D tracking with optical flow compensation
tracking, and extends both streams with multi-camera geometric verification to
maintain associations under rapid motion, initialization error, or local view
degradation.

\noindent\textbf{Stage 1: 3D-to-2D Tracking.} Let $\mathbf{T}_a^b$ denote the rigid transformation from frame $a$ to $b$. Using the pre-calibrated extrinsics $\{\mathbf{T}_{c_k}^{b}\}$ and the current body pose prediction $\mathbf{T}_{b_i}^{w}$, we project historical map points $\mathbf{P}_w^j \in \mathbb{R}^3$ into each camera $c_k$'s image plane through a unified multi-camera projection operator
\begin{equation}
\begin{aligned}
h_{c_k}(\mathbf{T}_{b}^{w}, \mathbf{P}_w^j) &\triangleq \pi_{c_k}\!\left(
\mathbf{T}_b^{c_k}(\mathbf{T}_b^w)^{-1}\mathbf{P}_w^j
\right)
\end{aligned}
\label{eq:hck}
\end{equation}
which is reused by tracking, factor-graph optimization, and global bundle
adjustment (BA) and relocalization. The Stage 1 prediction is then
\begin{equation}
\mathbf{p}_{c_k}^{j,\text{pred}} = h_{c_k}(\mathbf{T}_{b_i}^{w}, \mathbf{P}_w^j)
\end{equation}
where $\pi_{c_k} : \mathbb{R}^3 \to \mathbb{R}^2$ is the camera projection
function with distortion correction. Given the previous observation
$\mathbf{p}_{c_k}^{j,i-1}$, we find the precise match by performing local
optical flow matching within a neighborhood $\mathcal{N}_{c_k}^j$ around
$\mathbf{p}_{c_k}^{j,\text{pred}}$:
\begin{equation}
\begin{aligned}
\mathbf{p}_{c_k}^{j,*}
&= \mathbf{p}_{c_k}^{j,\text{pred}} + \arg \min_{\Delta \mathbf{p} \in
\Delta\mathcal{N}_{c_k}^j}
\sum_{\mathbf{u} \in \mathcal{W}} \\
&\quad \Big\| I_{c_k}^i(\mathbf{u} + \mathbf{p}_{c_k}^{j,\text{pred}} +
\Delta\mathbf{p}) - I_{c_k}^{i-1}(\mathbf{u} + \mathbf{p}_{c_k}^{j,i-1})
\Big\|^2
\end{aligned}
\label{eq:patch_refinement}
\end{equation}
where $\mathcal{W}$ is the local image patch and $\Delta\mathcal{N}_{c_k}^j$
denotes the local displacement search window. This pose-guided local search
narrows the matching window and improves robustness under fast motion.
Meanwhile, we support NN-Feature matching.

To further improve multi-camera tracking reliability, we introduce
cross-view geometric consistency verification. If the same map point
$\mathbf{P}_w^j$ is tracked in two distinct cameras $k_1$ and $k_2$ as
$(\mathbf{p}_{c_{k_1}}, \mathbf{p}_{c_{k_2}})$, their homogeneous coordinates
$(\tilde{\mathbf{p}}_{c_{k_1}}, \tilde{\mathbf{p}}_{c_{k_2}})$ must satisfy
the epipolar constraint defined by the relative pose
$\mathbf{T}_{c_{k_2}}^{c_{k_1}} = \mathbf{T}_b^{c_{k_1}}
(\mathbf{T}_b^{c_{k_2}})^{-1}$:
\begin{equation}
\begin{aligned}
(\tilde{\mathbf{p}}_{c_{k_1}})^T \mathbf{F}_{k_1,k_2}
\tilde{\mathbf{p}}_{c_{k_2}}
&= 0
\end{aligned}
\end{equation}
where $\mathbf{F}_{k_1,k_2}$ is the fundamental matrix. Matches with epipolar
residuals exceeding a threshold $\epsilon$ are flagged as outliers. The
cross-camera check removes associations that are inconsistent with the
calibrated rig geometry, which is especially useful when a single view suffers
from illumination changes, occlusion, or repetitive texture. Stage 1 outputs
$\mathcal{M}_{3D} = \{(\mathbf{P}_w^j, \mathbf{p}_{c_k}^{j,*})\}$.

\noindent\textbf{Stage 2: 2D-to-2D Tracking.} To accommodate newly extracted features lacking historical map points, as well as features where 3D-to-2D reprojection fails due to depth degradation from low texture, long range, or motion blur, we execute a pure visual frame-to-frame tracking fallback. Each camera $c_k$ independently runs feature association. The tracker parallelly utilizes both the local optical flow matching defined in (\ref{eq:patch_refinement}) and an independent NN-Feature matching modality. These distinct modalities generate candidate matches $(\mathbf{p}_{c_k}^{i-1,j}, \mathbf{p}_{c_k}^{i,j})$, which are then filtered by a lightweight geometric check within each camera. When one view provides too few reliable tracks, valid tracks from other cameras can still constrain the shared body motion, reducing the influence of a single degraded stream. The verification keeps geometrically consistent matches without coupling Stage 2 to a dense map. Stage 2 outputs $\mathcal{M}_{2D}$, which complements $\mathcal{M}_{3D}$ in low-depth or low-feature regions.

\subsection{Tightly Coupled Factor-Graph Optimization}
\label{subsec:fg}

In the multi-camera configuration, local BA parameterizes the sliding-window
optimization variables into a shared body state $\mathcal{X}$:
\begin{equation}
\begin{aligned}
\mathcal{X}_i
&= \begin{bmatrix}
\mathbf{T}_{b_i}^w & \mathbf{v}_{b_i}^w &
\mathbf{b}_{a_i} & \mathbf{b}_{g_i}
\end{bmatrix}
\end{aligned}
\end{equation}
where $\mathbf{T}_{b_i}^w = \begin{bmatrix} \mathbf{R}_{b_i}^w &
\mathbf{p}_{b_i}^w
\end{bmatrix} \in SE(3)$ is the body pose in the world frame,
$\mathbf{v}_{b_i}^w \in \mathbb{R}^3$ is the body velocity, and
$\mathbf{b}_{a_i}, \mathbf{b}_{g_i} \in \mathbb{R}^3$ are the IMU biases.

Rather than optimize each camera pose independently, the framework constrains
all visual and geometric observations to this shared body state with rigid
extrinsics $\{\mathbf{T}_{c_k}^{b} \in SE(3)\}$. This body state
parameterization keeps cross-view geometry consistent by construction and
avoids adding a separate pose variable for every camera.

\noindent\textbf{IMU Pre-integration Residual Factor $r_1$.} Inertial kinematic
constraints between adjacent keyframes $b_i$ and $b_{i+1}$ are formulated
through IMU pre-integration residuals. By integrating high-frequency IMU in the local body state, we avoid the
computational overhead of re-integration during iterative state updates. The
residual vector on the inertial state error manifold is:
\begin{equation}
\begin{aligned}
\mathbf{r}_1
&= \mathbf{r}_{imu}(\mathbf{z}_{b_i,b_{i+1}}, \mathcal{X}_i,
\mathcal{X}_{i+1}) \\
&= \begin{bmatrix}
\mathbf{r}_{\Delta R} \\ \mathbf{r}_{\Delta v} \\
\mathbf{r}_{\Delta p} \\ \mathbf{r}_{b_a} \\ \mathbf{r}_{b_g}
\end{bmatrix}
\end{aligned}
\end{equation}
with the explicit nonlinear prediction errors:
\begin{equation}
\begin{aligned}
\mathbf{r}_{\Delta R} &= \log \Big( \big( \Delta \tilde{\mathbf{R}}_{b_i
b_{i+1}} \exp ( \mathbf{J}_{b_g}^R \delta \mathbf{b}_{g_i} ) \big)^{-1} \\
&\quad \cdot (\mathbf{R}_{b_i}^w)^{-1} \mathbf{R}_{b_{i+1}}^w \Big), \\
\mathbf{r}_{\Delta v} &= (\mathbf{R}_{b_i}^w)^{-1} \left( \mathbf{v}_{b_{i+1}}^w
- \mathbf{v}_{b_i}^w - \mathbf{g}^w \Delta t \right) \\ &\quad - \left( \Delta
\tilde{\mathbf{v}}_{b_i b_{i+1}} + \mathbf{J}_{b_a}^v
\delta \mathbf{b}_{a_i} + \mathbf{J}_{b_g}^v \delta \mathbf{b}_{g_i} \right), \\
\mathbf{r}_{\Delta p} &= (\mathbf{R}_{b_i}^w)^{-1} \left( \mathbf{p}_{b_{i+1}}^w
- \mathbf{p}_{b_i}^w - \mathbf{v}_{b_i}^w \Delta t - \tfrac{1}{2}\mathbf{g}^w
\Delta t^2 \right) \\
&\quad - \left( \Delta \tilde{\mathbf{p}}_{b_i b_{i+1}} + \mathbf{J}_{b_a}^p
\delta \mathbf{b}_{a_i} + \mathbf{J}_{b_g}^p \delta \mathbf{b}_{g_i} \right)
\end{aligned}
\end{equation}
where $\{\Delta \tilde{\mathbf{R}}, \Delta \tilde{\mathbf{v}}, \Delta
\tilde{\mathbf{p}}\}$ are the pre-integrated measurements,
$\mathbf{J}_{(\cdot)}^{(\cdot)}$ are the first-order Jacobian corrections with
respect to the biases, and $\mathbf{g}^w$ is the gravity vector in the world
frame. The random-walk bias residuals are
$\mathbf{r}_{b_a}=\mathbf{b}_{a_{i+1}}-\mathbf{b}_{a_i}$ and
$\mathbf{r}_{b_g}=\mathbf{b}_{g_{i+1}}-\mathbf{b}_{g_i}$. The Mahalanobis
weight of the IMU factor is dictated by the covariance matrix $\Sigma_{imu}$,
propagated by the error-state transition law. The resulting factor provides
inertial scale information and short-term trajectory continuity constraints.

\noindent\textbf{Multi-Camera Visual Reprojection Residual Factor $r_2$.} We
fuse synchronized visual observations from all $N$ cameras. This visual factor
forms the base geometric term, while depth-related weighting is applied when
valid depth observations are available from either RGB-D sensor or the
optional pseudo-depth interface. At keyframe $i$, a map point
$\mathbf{P}_w^j$ projects onto the $k$-th camera's image plane via:
\begin{equation}
\mathbf{p}_{c_k}^{j,\text{proj}} = h_{c_k}(\mathbf{T}_{b_i}^{w}, \mathbf{P}_w^j)
\end{equation}
where $h_{c_k}$ is the unified projection operator from Eq.~\eqref{eq:hck},
with $\mathbf{T}_b^{c_k} = (\mathbf{T}_{c_k}^b)^{-1}$. The unified visual cost
is:
\begin{equation}
\begin{aligned}
r_2(i) &= \sum_{k=0}^{N-1}
\sum_{j \in \mathcal{F}_{i,k}}
\rho_{H}\!\left(
\left\|\mathbf{p}_{i,k}^{j}
- \mathbf{p}_{c_k}^{j,\text{proj}}
\right\|_{\Sigma_{v,j}}^2
\right)
\end{aligned}
\end{equation}
where $\mathcal{F}_{i,k}$ is the set of visual observations in camera $c_k$ at
keyframe $i$, $\mathbf{p}_{i,k}^{j}$ is the measured pixel coordinate, and
$\rho_{H}(\cdot)$ is the Huber robust kernel. The measurement covariance
$\Sigma_{v,j}$ is adaptively weighted according to the feature pyramid level
and the validity of the associated depth observation when the depth module is
enabled. The visual residual links multi-camera observations to the global body
state and keeps all camera measurements in one optimization frame.

\noindent\textbf{GICP Relative Pose Residual Factor $r_3$.} The GICP factor is
depth-gated: it is instantiated only when valid depth observations are
available. These observations may come from physical RGB-D sensors or from the
optional pseudo-depth interface. In textureless corridors or environments
with drastic illumination transitions where visual features degrade severely,
the resulting local point clouds provide supplementary GICP constraints.
Following the GICP penalizes both point-to-point Euclidean distances and probabilistic local
surface structures. Let $d_i$ and $d_j$ denote the keyframes $i$ and $j$. The
GICP factor is:
\begin{equation}
\begin{aligned}
r_3 &= \sum
\rho_{C}\!\left(
\left\|\mathbf{T}_{i,j}\right\|_{
\Sigma_{\text{GICP}, i,j}}^2
\right)
\end{aligned}
\end{equation}
where $ {T}_{i,j} $ is the relative pose of GICP and $\rho_{C}(\cdot)$
denotes the robust Cauchy kernel.
\subsection{Loop Closure and Global BA}
\label{subsec:loop_closing}
To reduce accumulated drift and scale divergence in large-scale trajectories,
GeoFlow-SLAM++ uses a multi-camera loop closure and global factor-graph
optimization mechanism.

\begin{figure}[!tbp]
\centering
\includegraphics[width=\columnwidth]{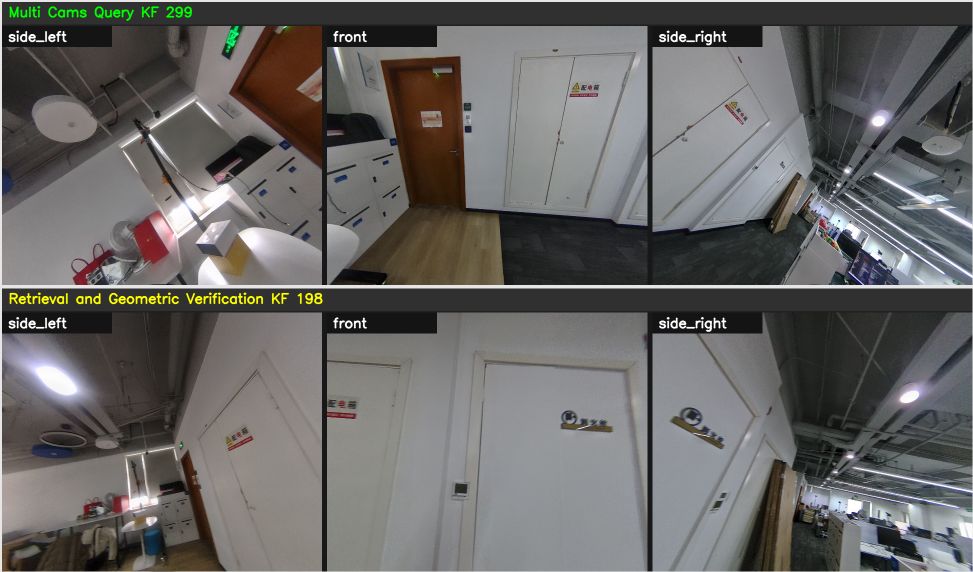}
\caption{\textbf{Unified cross-view place recognition retrieval.} A query keyframe at time $t_i$ provides left-side, front, and right-side views, and the retrieved candidate at time $t_j$ has complementary overlap across the same camera system. GeoFlow-SLAM++ aggregates per-camera descriptors into $\mathbf{v}_{\text{unified}}$ and queries the map database in the unified visual descriptor space. The retrieved candidate is then used for geometric verification, loop closure, and cross-session relocalization.}
\label{fig:unified_pr}
\end{figure}

\noindent\textbf{Unified Cross-View Place Recognition.} Conventional multi-camera
SLAM frameworks typically perform loop closure retrieval on a per-camera basis. This independent retrieval strategy tends to produce false negative results when the robot undergoes aggressive maneuvers, partial occlusion occurs in a single view, or drastic illumination fluctuations affect individual sensors. 
GeoFlow-SLAM++ extends appearance-based loop closure to a multi-camera fused place recognition module, which can run seamlessly with either of the two feature front-ends. Whenever a keyframe is created, global descriptors $\mathbf{v}_{c_k}$ are extracted in parallel from all active
cameras $\{c_k\}_{k=0}^{N-1}$. Since all descriptors belong to the same
calibrated body keyframe, they are combined into the unified descriptor
$\mathbf{v}_{\text{unified}}$ by $\mathcal{A}(\cdot)$. The similarity query is
then executed against the map database in the corresponding descriptor space.
When some camera views suffer degradation, the remaining views can still
contribute appearance evidence for retrieval in complex environments.
Fig.~\ref{fig:unified_pr} illustrates how complementary camera views are
aggregated for retrieval under partial view degradation.

\noindent\textbf{Global BA.} Once a loop candidate passes
 geometric verification in cross-view, global BA uses the loop constraint to
update the trajectory and map jointly. Unlike the local sliding-window
optimization, global BA revisits all historical body states, IMU biases, and
map points, and jointly balances inertial, visual, and GICP constraints:
\begin{equation}
\begin{aligned}
\min_{\mathcal{X}, \mathbf{P}_w} \;\;
&\sum \rho_{I}({r}_{\text{imu}}) + \sum \rho_{H}({r}_{\text{vis}}) \\ &\quad +
\sum \rho_{C}({r}_{\text{GICP}})
\end{aligned}
\end{equation}
where the visual term uses the multi-camera reprojection residual
\begin{equation}
\begin{aligned}
\mathbf{e}_{i,k,j}
&= \mathbf{p}_{i,k}^{j} - h_{c_k}(\mathbf{T}_{b_i}^{w}, \mathbf{P}_w^j)
\end{aligned}
\end{equation}
and the other terms follow the same factor definitions as the local factor
graph. Solving this sparse least-squares problem distributes the loop
correction over the trajectory.

\subsection{Relocalization}
\label{subsec:geometric_reloc}
GeoFlow-SLAM++ uses the same multi-camera aggregation strategy for cross-session
relocalization. When relocalization is triggered, the system extracts global descriptors $\mathbf{v}_{c_k}$ and aggregates them into
$\mathbf{v}_{\text{unified}}$ to query the map database, \begin{equation}
\mathbf{v}_{\text{unified}} = \mathcal{A}(\{\mathbf{v}_{c_k}\}_{k=0}^{N-1}).
\end{equation}

For the ORB front-end, $\mathcal{A}(\cdot)$ corresponds to summing sparse BoW
vectors; for the NN-Feature front-end, it denotes normalized pooling of
per-camera descriptors. Querying $\mathbf{v}_{\text{unified}}$
against the inverted index of the prior map, together with cumulative scoring
over covisible keyframes, yields candidate matches. When some cameras are
occluded or visually degraded, the remaining views can still contribute
appearance evidence for relocalization.

For retrieved candidates, a joint multi-camera Perspective-n-Point (PnP)
solver estimates absolute pose constraints directly on the body state's
$SE(3)$ manifold, reducing the need for separate per-camera pose
initializations. With rigid extrinsics $\{\mathbf{T}_{c_k}^b\}$, 2D
observations from all cameras and their corresponding map points
$\mathbf{P}_w^j$ are jointly constrained by the multi-camera reprojection
model:
\begin{equation}
\mathbf{p}_{c_k}^j = h_{c_k}(\mathbf{T}_{b}^{w}, \mathbf{P}_w^j)
\end{equation}
where $\mathbf{T}_b^w$ is the target body pose. Pooling matches across cameras
allows random sample consensus (RANSAC) hypotheses to draw support from the
cross-view inlier set rather than from one camera alone.

After initial pose estimation, coarse-to-fine geometric checks and motion-only
BA filter out spurious matches across all cameras. The system accepts a
relocalization hypothesis when the verified multi-view support is sufficient:
\begin{equation}
\begin{aligned}
n_{\text{inlier}} &= \sum_{k=0}^{N-1} \sum_{j \in \mathcal{I}_{k}} 1
\;\ge\; \gamma_{\text{thresh}}.
\end{aligned}
\end{equation}
where $\mathcal{I}_{k}$ is the verified inlier set for camera $c_k$. This
criterion prevents low-support hypotheses from entering localization mode.

After relocalization, the system enters pure localization mode: local mapping
and loop closure stop, while tracking continues against a local submap cropped
from the Map with IMU propagation and joint multi-camera reprojection.

\subsection{Optional Multi-Camera Joint Feed-Forward Depth Interface}
\label{subsec:depth}
The depth module is an optional source of supplementary geometry. When
enabled, it infers dense pseudo-depth from synchronized RGB images and
fixed multi-camera extrinsics. At time step $i$, the system uses images from
$N$ cameras:
\begin{equation}
\mathcal{I}_i = \{I_{c_k}^i\}_{k=0}^{N-1}, \quad N \geq 3
\end{equation}
alongside fixed rigid extrinsics relative to the body:
\begin{equation}
\mathcal{T}_c = \{\mathbf{T}_{c_k}^{b}\}_{k=0}^{N-1}.
\end{equation}

A joint pseudo-depth model predicts all camera depths in one pass:
\begin{equation}
\begin{aligned}
\{D_{c_k}^i\}_{k=0}^{N-1}
&= \Phi_\theta\big(\{I_{c_k}^i\}_{k=0}^{N-1}, \\ &\quad
\{\mathbf{T}_{c_k}^{b}\}_{k=0}^{N-1}\big)
\end{aligned}
\end{equation}
where $\Phi_\theta$ is the joint pseudo-depth model and $D_{c_k}^i$ is the
pseudo-depth of the $k$-th camera at time step $i$.

\noindent\textbf{Unified Structural Alignment.} Because the pseudo-depths are tied to the
same calibrated multi-camera system, each depth can be transformed into a
common reference frame for map recovery and geometric residuals:
\begin{equation}
\begin{aligned}
\mathbf{P}_{c_0}(\mathbf{p}_{c_k})
&= \mathbf{T}_{c_k}^{c_0}\,
\pi_{c_k}^{-1}\big(\mathbf{p}_{c_k}, D_{c_k}(\mathbf{p}_{c_k})\big)
\end{aligned}
\end{equation}
where $\pi_{c_k}^{-1}(\cdot)$ back-projects pixel $\mathbf{p}_{c_k}$ with its
pseudo-depth and $\mathbf{T}_{c_k}^{c_0}$ maps camera $c_k$ to reference
camera $c_0$. To limit the effect of pseudo-depth errors,
cross-view-consistent depth observations enter map updates and residual
construction.

\section{Experimental Evaluation}
\label{sec:experiments}

\begin{table*}[ht]
  \caption{Quantitative Comparison of Localization Accuracy on the Hilti
Benchmark without Pseudo-Depth. Methods are grouped into LiDAR-assisted
systems and vision-only visual-inertial systems. ATE RMSE is measured in meters. ``$-$'' indicates tracking failure; the average row is
reported only for methods that successfully complete all six sequences to
ensure a fair comparison. Best is in \textbf{bold}, second-best
\underline{underlined}.}
  \label{tab:hilti_mapping}
  \centering
  \resizebox{\textwidth}{!}{
  \begin{tabular}{l@{\hspace{4mm}}cccccccc}
  \toprule
  \multirow{2}{*}{\textbf{Sequence}} &
  \multicolumn{5}{c}{\textbf{LiDAR-assisted systems}} &
  \multicolumn{3}{c}{\textbf{Vision-only visual-inertial systems}} \\
  \cmidrule(r){2-6} \cmidrule(l){7-9}
& \textbf{FAST-LIVO} & \textbf{R3LIVE} & \textbf{FAST-LIO2} &
  \textbf{FAST-LIVO2} & \textbf{Omni-LIVO} & \textbf{OpenMAVIS} &
  \textbf{\makecell{GeoFlow-SLAM++\\(ORB)}} &
  \textbf{\makecell{GeoFlow-SLAM++\\(NN-Feature)}} \\ 
  \midrule
Exp04 & 0.032 & \underline{0.028} & 0.124 & 0.030 & \textbf{0.027} & 0.083 &
0.036 & 0.048 \\ Exp05 & 0.089 & 0.041 & 0.100 & 0.041 & \textbf{0.009} &
\underline{0.040} & 0.041 & 0.047 \\ Exp06 & 0.091 & 0.049 & $-$    &
\underline{0.045} & \textbf{0.011} & 0.075 & 0.047 & 0.081 \\ Exp14 & 0.129 &
0.097 & 0.136 & 0.097 & \underline{0.031} & 0.082 & 0.034 &
  \textbf{0.030} \\
Exp16 & 1.165 & $-$    & $-$    & 0.123 & 0.120 & 0.255 & \underline{0.101} &
  \textbf{0.091} \\
Exp18 & 0.387 & $-$    & $-$    & 0.155 & \textbf{0.083} & 0.178 & 0.091 &
  \underline{0.089} \\
  \midrule
  \textbf{Avg.} & 0.316 & $-$ & $-$ & 0.082 & \textbf{0.047} & 0.119 & \underline{0.058} & 0.064 \\
  \bottomrule
  \end{tabular}
  }
\end{table*}

\begin{figure*}[ht]
  \centering
  \includegraphics[width=0.90\textwidth]{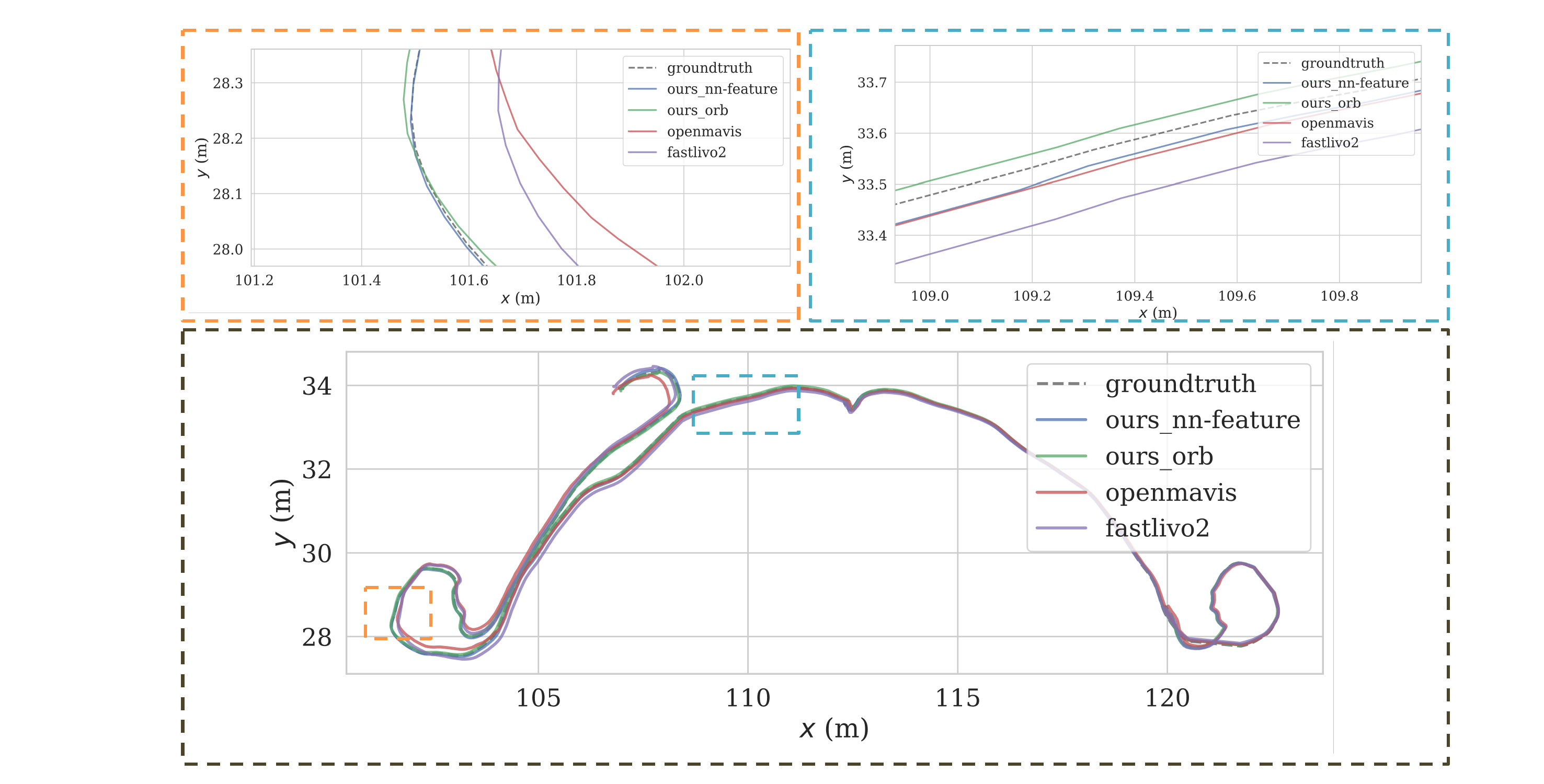}
  \vspace{-0.6em}
  \caption{Qualitative mapping trajectory comparison with representative
baselines on Hilti Exp18. The plotted trajectories are aligned to the ground
truth for visualization. The zoomed regions compare local drift with
FAST-LIVO2 and OpenMAVIS, which represent LiDAR-visual-inertial and
multi-camera visual-inertial baselines, respectively.}
  \label{fig:hilti_mapping_traj}
\end{figure*}

We comprehensively benchmark GeoFlow-SLAM++ across five distinct datasets from four evaluation dimensions. The Hilti SLAM Challenge dataset~\cite{hilti} facilitates quantitative comparisons between the presented multi-camera RGB-IMU system and mainstream LiDAR-based SLAM pipelines, with the pseudo-depth module disabled throughout evaluations. A self-collected \textit{Pocket2} dataset, recorded via the MANIFOLD TECH handheld Pocket2 device with three fisheye cameras and a Livox Mid360 LiDAR, is used to evaluate pure vision-based cross-session relocalization. Ablation studies on the TUM RGB-D~\cite{tumrgbd} and Hilti datasets systematically quantify the individual contributions of loop closure, optical flow tracking, NN-Feature, and the optional depth branch. Additionally, EuRoC~\cite{euroc} and OpenLORIS~\cite{openloris} further validate the efficacy of NN-Feature for robust RGB-IMU and RGB-D tracking under challenging conditions, including texture deficiency, complex appearance variations, and drastic illumination fluctuations.

\noindent\textbf{Evaluation Details.} Depth Anything V3 \cite{depthanythingv3} is exclusively adopted for ablation experiments involving the depth module, where all depth predictions are precomputed offline. The GICP constraint is activated adaptively whenever valid RGB-D measurements or reliable pseudo-depth observations are available.

\subsection{Multi-Camera Mapping Evaluation on Hilti Benchmark}\label{subsec:exp_mapping} Table~\ref{tab:hilti_mapping} summarizes the localization accuracy on the Hilti, where the root mean square error (RMSE) of Absolute Trajectory Error (ATE) is taken as the quantitative evaluation metric. GeoFlow-SLAM++(ORB) achieves an average ATE of $0.058$~m, significantly outperforming OpenMAVIS. While Omni-LIVO yields the highest overall accuracy, our ORB-based variant remains competitive, demonstrating the effectiveness of our visual-only approach against LiDAR-assisted baselines.

The superiority of our framework is most evident in challenging sequences Exp14, Exp16, and Exp18. In these scenarios, GeoFlow-SLAM++(NN-Feature) effectively overcomes poor visual conditions, outperforming Omni-LIVO by 3.23\% and 24.17\% on Exp14 and Exp16, respectively. Notably, several LiDAR-assisted baselines fail to complete these sequences, whereas our system maintains consistent tracking. In geometrically regular scenes Exp04 through Exp06, our variants show only minor deviations from the Omni-LIVO benchmark, confirming their high precision.

The slightly higher average ATE of $0.064$~m for the NN-Feature variant is attributed to the fact that ORB features are sufficient for stable matching in structured scenes within the feature budget. As illustrated in Fig.~\ref{fig:hilti_mapping_traj}, trajectory comparisons in Exp18 further verify that both variants effectively suppress cumulative drift compared to FAST-LIVO2 and OpenMAVIS.

\begin{table*}[ht]
  \caption{Quantitative Evaluation of Relocalization and Localization Accuracy and
Efficiency in Cross-Session Scenarios Using the ORB Front-End. RMSE in meters
and relocalization time in milliseconds. Best 6-DoF
RMSE is \underline{underlined}, and best 7-DoF RMSE is in \textbf{bold}.}
  \label{tab:relocalization}
  \centering
  \begin{tabular}{l@{\hspace{4mm}}ccc@{\hspace{3mm}}ccc@{\hspace{3mm}}ccc}
  \toprule
  \multirow{2}{*}[-0.3ex]{\textbf{Scene}} & \multirow{2}{*}[-0.3ex]{\textbf{\makecell{Mapping\\Sequence}}} &
  \multirow{2}{*}[-0.3ex]{\textbf{\makecell{Localization\\Sequence}}} & \multirow{2}{*}[-0.3ex]{\textbf{\makecell{Total\\Frames}}} &
  \multicolumn{3}{c}{\textbf{LiDAR-Based}} & \multicolumn{3}{c}{\textbf{GeoFlow-SLAM++ (ORB)}} \\
  \cmidrule(r){5-7} \cmidrule(l){8-10}
& & & & \textbf{6-DoF} & \textbf{7-DoF} & \textbf{Time} & \textbf{6-DoF} &
\textbf{7-DoF} & \textbf{Time} \\
  \midrule
Sh3    & 20260320-1309 & 20260414-1242 & 1288 & 0.077 & 0.078 & 87.660 &
\underline{0.048} & \textbf{0.015} & 99.050 \\ Sh2    & 20260414-1340 &
20260320-1315 & 2255 & 0.087 & 0.087 & 91.180 & \underline{0.054} &
\textbf{0.036} & 98.540 \\ Sh1    & 20260422-1040 & 20260414-1345 & 1583 &
\underline{0.072} & 0.069 & 84.080 & 0.145 & \textbf{0.034} & 87.530 \\ Garage
& 20260422-1422 & 20260422-1426 & 2405 & \underline{0.082} & \textbf{0.078} &
90.260 & 0.122 & 0.086 & 99.360 \\
  \midrule
  \textbf{Avg.} & $-$ & $-$ & $-$ & \underline{0.079} & 0.078 & 88.290 & 0.092 & \textbf{0.043} & 96.120 \\
  \bottomrule
  \end{tabular}
\end{table*}

\begin{table*}[ht]
  \caption{Ablation Study Results of the NN-Feature Front-End on the TUM RGB-D
Dataset. ``Op'' denotes dual-stream optical flow tracking. ATE RMSE in meters.
``$-$'' indicates tracking failure. Best is in \textbf{bold},
second-best \underline{underlined}.}
  \label{tab:ablation_rgbd}
  \centering
  \begin{tabular}{ccc@{\hspace{4mm}}ccccccccc@{\hspace{4mm}}c}
  \toprule
  \textbf{Loop} & \textbf{Op} & \textbf{NN-Feature} & \textbf{360} & \textbf{desk} &
  \textbf{desk2} & \textbf{floor} & \textbf{plant} & \textbf{room} & \textbf{rpy}
& \textbf{teddy} & \textbf{xyz} & \textbf{Avg.} \\
  \midrule
$\times$ & $\times$ & $\times$ & 0.153 & 0.022 & 0.305 & - & 0.050 & 0.338 &
0.023 & 0.195 & \underline{0.011} & 0.137 \\ $\times$ & $\checkmark$ &
$\times$ & 0.155 & 0.026 & 0.157 & 0.222 & 0.117 & 0.250 & \underline{0.021} &
0.254 & \underline{0.011} & 0.135 \\ $\times$ & $\times$ & $\checkmark$ &
\underline{0.094} & 0.019 & \underline{0.044} & \underline{0.034} & 0.034 &
0.193 & 0.022 & \underline{0.144} & \textbf{0.010} & \underline{0.066} \\
$\checkmark$ & $\times$ & $\times$ & 0.176 & 0.018 & 0.209 & - &
\underline{0.028} & 0.201 & \textbf{0.020} & 0.170 & \underline{0.011} & 0.104
\\ $\checkmark$ & $\checkmark$ & $\times$ & 0.120 & \underline{0.017} &
\textbf{0.026} & 0.287 & \textbf{0.018} & \textbf{0.078} & \underline{0.021} &
0.163 & \textbf{0.010} & 0.082 \\ $\checkmark$ & $\times$ & $\checkmark$ &
\textbf{0.081} & \textbf{0.016} & 0.051 & \textbf{0.017} & 0.033 &
\underline{0.090} & 0.024 & \textbf{0.128} & \textbf{0.010} &
  \textbf{0.050} \\
  \bottomrule
  \end{tabular}
\end{table*}

\begin{table*}[ht]
  \caption{Ablation Study Results with Optional Pseudo-Depth on the Hilti
Multi-Camera Dataset. ``Op'' denotes dual-stream optical flow tracking. ATE
RMSE in meters. Best is in \textbf{bold}, second-best
\underline{underlined}.}
  \label{tab:ablation_hilti}
  \centering
  \begin{tabular}{cccc@{\hspace{4mm}}ccccccc}
  \toprule
  \textbf{Loop} & \textbf{Op} & \textbf{NN-Feature} & \textbf{Depth} & \textbf{Exp04} &
  \textbf{Exp05} & \textbf{Exp06} & \textbf{Exp14} & \textbf{Exp16} &
  \textbf{Exp18} & \textbf{Avg.} \\
  \midrule
$\times$ & $\times$ & $\times$ & $\times$ & 0.055 & 0.042 & 0.068 & 0.038 &
0.113 & 0.107 & 0.070 \\ $\checkmark$ & $\times$ & $\times$ & $\times$ & 0.040
& 0.040 & 0.064 & 0.039 & 0.105 & 0.104 & 0.065 \\ $\times$ & $\checkmark$ &
$\times$ & $\times$ & \underline{0.038} & \textbf{0.027} &
  \underline{0.049} & 0.040 & 0.122 & 0.094 & 0.062 \\
$\checkmark$ & $\checkmark$ & $\times$ & $\times$ & \textbf{0.036} & 0.042 &
  \textbf{0.048} & 0.034 & 0.102 & \underline{0.091} & \textbf{0.059} \\
$\times$ & $\times$ & $\checkmark$ & $\times$ & 0.041 & 0.059 & 0.086 &
  \textbf{0.029} & \textbf{0.087} & 0.092 & 0.066 \\
$\checkmark$ & $\times$ & $\checkmark$ & $\times$ & 0.048 & 0.047 & 0.081 &
  \underline{0.030} & \underline{0.091} & \textbf{0.089} & 0.064 \\
$\checkmark$ & $\times$ & $\checkmark$ & $\checkmark$ & 0.039 &
\underline{0.034} & \textbf{0.048} & 0.037 & 0.107 & 0.092 & \underline{0.060}
\\
  \bottomrule
  \end{tabular}
\end{table*}

\subsection{Cross-Session Relocalization}
\label{subsec:exp_relocalization}
We evaluate cross-session relocalization under practical vision-only deployment conditions, where localization relies solely on fisheye cameras without LiDAR inputs during runtime.
We conduct experiments on four self-collected sequences, including three indoor scenes with appearance variations and one low-texture parking scenario. For each scene, one sequence is used for map construction and another cross-session sequence serves as the localization query, with LiDAR-based localization adopted as the benchmark.
We adopt six-degree-of-freedom (6-DoF) and seven-degree-of-freedom (7-DoF) RMSE for comprehensive evaluation, which quantify global registration accuracy and trajectory intrinsic consistency, respectively.

As shown in Table~\ref{tab:relocalization}, our GeoFlow-SLAM++ achieves superior trajectory stability. It achieves an average 7-DoF RMSE of 0.043 m, reducing the LiDAR baseline error by 44.87\% and outperforming the LiDAR method on all three indoor sequences. The LiDAR baseline achieves better global alignment in terms of 6-DoF accuracy, since visual matching is susceptible to illumination changes, object rearrangement and low texture across different sessions. In terms of efficiency, our vision-only method achieves comparable runtime with the LiDAR baseline, demonstrating good practicality for real-world deployment.

\begin{table*}[ht]
  \centering
  \caption{Quantitative Comparison of Monocular RGB-IMU Localization Accuracy of GeoFlow-SLAM++ (NN-Feature) on the EuRoC Dataset. ATE RMSE in meters. ``$-$'' indicates tracking failure. Best is in \textbf{bold}, second-best \underline{underlined}.}
  \label{tab:euroc_mono}
  \begingroup
  \small
  \setlength{\tabcolsep}{3pt}
  \renewcommand{\arraystretch}{0.9}
  \begin{tabular}{l@{\hspace{4mm}}ccccccc}
  \toprule
  \textbf{Dataset} & \textbf{ORB-SLAM3} & \textbf{OKVIS} & \textbf{VINS-Mono-loop} & \textbf{VINS-Fusion-loop} & \textbf{SuperVINS} & \textbf{GeoFlow-SLAM} & \textbf{\makecell{GeoFlow-SLAM++\\(NN-Feature)}} \\
  \midrule
MH01 & 0.062 & 0.330 & 0.087 & 0.091 & 0.087 & \textbf{0.019} &
\underline{0.025} \\ MH02 & 0.037 & 0.370 & 0.108 & 0.046 & 0.097 &
\underline{0.035} & \textbf{0.027} \\ MH03 & \underline{0.046} & 0.250 & 0.066
& 0.092 & 0.181 & 0.052 & \textbf{0.038} \\ MH04 & 0.075 & 0.270 & 0.205 &
0.131 & 0.171 & \underline{0.071} & \textbf{0.069} \\ MH05 & \textbf{0.057} &
0.390 & 0.142 & 0.262 & 0.158 & \underline{0.070} & 0.100 \\ V101 & 0.049 &
0.094 & 0.048 & 0.154 & 0.201 & \underline{0.034} & \textbf{0.031} \\ V102 &
\textbf{0.015} & 0.140 & 0.063 & 0.219 & 0.116 & \underline{0.027} & 0.028 \\
V103 & \textbf{0.037} & 0.210 & 0.201 & 0.170 & 0.221 & 0.096 &
\underline{0.069} \\ V201 & \underline{0.042} & 0.090 & 0.067 & 0.123 & 0.061
& \textbf{0.031} & 0.046 \\ V202 & \underline{0.021} & 0.170 & 0.159 & $-$
& 0.100 & \textbf{0.018} & 0.025 \\ V203 & \textbf{0.027} & 0.230 & 0.248 &
0.193 & 0.169 & \underline{0.029} & 0.093 \\
  \midrule
  \textbf{Avg.} & \textbf{0.043} & 0.231 & 0.127 & 0.148 & 0.142 & \underline{0.044} & 0.050 \\
  \bottomrule
  \end{tabular}
  \endgroup

  \vspace{1.5em} 

  \caption{Quantitative Comparison of Monocular RGB-D-IMU Localization Accuracy of GeoFlow-SLAM++ (NN-Feature) on the OpenLORIS-Scene Dataset. ATE RMSE in meters. ``$-$'' indicates tracking failure. Best is in \textbf{bold}, second-best \underline{underlined}.}
  \label{tab:openloris_rgbd}
  \begingroup
  \small
  \setlength{\tabcolsep}{6pt} 
  \renewcommand{\arraystretch}{0.9}
  \begin{tabular}{lcccccc}
  \toprule
  \textbf{Dataset} & \textbf{ORB-SLAM3} & \textbf{VINS-Mono} & \textbf{VINS-RGBD} & \textbf{S-VIO} & \textbf{GeoFlow-SLAM} & \textbf{\makecell{GeoFlow-SLAM++\\(NN-Feature)}} \\
  \midrule
Home1-1   & 0.929 & 1.076 & 0.452 & \underline{0.402} & 0.458 & \textbf{0.339}
\\ Home1-2   & $-$   & 0.542 & 0.379 & \textbf{0.316} & \underline{0.378} &
0.381 \\ Office1-1 & 0.068 & 0.231 & 0.101 & 0.101 & \underline{0.058} &
\textbf{0.056} \\ Office1-2 & 0.123 & 0.289 & 0.124 & 0.099 &
\underline{0.070} & \textbf{0.069} \\ Office1-3 & $-$   & 0.155 & 0.154 &
0.151 & \textbf{0.029} & \underline{0.100} \\ Office1-4 & 0.253 & 0.392 &
0.186 & 0.164 & \underline{0.155} & \textbf{0.148} \\ Office1-5 & $-$   &
0.238 & 0.239 & 0.212 & \underline{0.200} & \textbf{0.197} \\
  \midrule
  \textbf{Avg.} & $-$ & 0.418 & 0.234 & 0.206 & \underline{0.193} & \textbf{0.184} \\
  \bottomrule
  \end{tabular}
  \endgroup
\end{table*}

\subsection{Ablation Study}
\label{subsec:ablation_study}

We conduct ablation experiments to quantify the individual contribution of loop closure (\textbf{Loop}), dual-stream optical flow tracking (\textbf{Op}), NN-Feature and pseudo-depth, as well as the complementarity among these modules.
\subsubsection{TUM RGB-D}
As shown in Table~\ref{tab:ablation_rgbd}, all the modules designed in this paper can improve localization accuracy to varying degrees. When used alone, the \textbf{NN-Feature} module achieves an average ATE of $0.066$~m, which is clearly better than the standalone \textbf{Loop} and \textbf{Op} modules. This module provides the most substantial performance improvement: it not only restores effective tracking on the floor sequence where the baseline loses track, but also significantly suppresses severe cumulative pose drift in the challenging desk2 sequence. \textbf{Loop} mainly constrains long-term global localization drift and achieves the most obvious accuracy improvement on the long-trajectory room sequence. The performance gain brought by standalone \textbf{Op} is quite limited, which indicates that for such short sequences constrained by metric depth, optical flow is mainly adopted to stabilize short-term frame-to-frame tracking rather than eliminate global drift. The combination of \textbf{Loop} and \textbf{NN-Feature} achieves the best average ATE of $0.050$~m, indicating clear complementarity between the two modules, and it outperforms all single-module schemes as well as the \textbf{Loop}+\textbf{Op} combination.

\subsubsection{Multi-Camera Hilti}
As shown in Table~\ref{tab:ablation_hilti}, all methods achieve centimeter-level localization accuracy with negligible performance gaps under the multi-camera setting. Compared with the TUM datasets, each module exhibits different performance preferences. Benefiting from sufficient visual redundancy of multi-camera systems, the baseline can maintain stable tracking, and each module only brings marginal accuracy gains. The \textbf{Loop}+\textbf{Op} combination achieves the lowest average ATE of $0.059$~m. The \textbf{NN-Feature} module can effectively restrict cumulative pose drift on several test sequences, while it performs worse than \textbf{Op} on other sequences. Integrating pseudo-depth into the \textbf{Loop}+\textbf{NN-Feature} pipeline strengthens geometric constraints, yet depth estimation noise slightly reduces localization accuracy in several dynamic scenarios. The full pipeline with pseudo-depth achieves an average ATE of $0.060$~m, which is comparable to the optimal scheme. This suggests that pseudo-depth provides geometric compensation only in limited scenarios and does not further improve global localization accuracy in stable multi-camera SLAM systems with abundant visual redundancy.

\subsection{Experiments on EuRoC and OpenLORIS}
\label{subsec:euroc_openloris}
To isolate the contribution of the dual front-end design, we further evaluate the NN-Feature branch under monocular RGB-IMU EuRoC and monocular RGB-D OpenLORIS settings, decoupled from the full multi-camera system. These experiments are intended to verify the standalone benefit of the learning-based front-end within the overall GeoFlow-SLAM++ design.
\subsubsection{EuRoC} 
As shown in Table~\ref{tab:euroc_mono}, GeoFlow-SLAM++ (NN-Feature) achieves the lowest ATE on sequences MH02, MH03, MH04, and V101, demonstrating that it can deliver more stable feature associations under scenarios with viewpoint changes, motion blur, and weak textures. Its performance degrades on sequences V201 to V203, which is mainly attributed to unstable long-baseline feature tracking under high-speed motion. The overall average ATE reaches \(0.050\)~m, yielding comparable localization accuracy against conventional approaches.
\subsubsection{OpenLORIS} As summarized in Table~\ref{tab:openloris_rgbd}, GeoFlow-SLAM++ (NN-Feature) achieves the best results on Home1-1, Office1-1, Office1-2, Office1-4, and Office1-5, with the minimum average ATE of \(0.184\)~m. Although depth information is incorporated into the system to provide scale and geometric constraints, NN-Feature further reduces localization errors by enhancing the stability of feature matching. This reveals that high-quality visual correspondences are still essential for boosting SLAM localization precision.

\section{Conclusion}
\label{sec:conclusion}
In this study, we present GeoFlow-SLAM++, a robust and tightly coupled multi-camera VI-SLAM framework that extends our prior single-sensor GeoFlow-SLAM into a unified body-centric multi-camera system. A central design of GeoFlow-SLAM++ is its dual visual front-end design, which supports both conventional ORB-based tracking and learning-based NN-Feature tracking under a shared downstream geometric pipeline. Built on this formulation, the system integrates multi-camera dual-stream optical flow/NN-Feature tracking, cross-view place recognition, and tightly coupled factor-graph optimization for mapping and relocalization. An optional pseudo-depth extension further provides supplementary geometric constraints in selected degraded scenarios. GeoFlow-SLAM++ achieves competitive performance across multiple benchmark datasets.

Beyond straightforward pseudo-depth integration in geometric SLAM, our future work will focus on improving the geometric consistency of depth estimated by 3D foundation models, aligning the scale of pseudo-depth with sensor measurements, and investigating whether such pseudo-depth can replace physical depth sensors for high-precision mapping and localization on robotic platforms.
\FloatBarrier

\bibliographystyle{IEEEtran}
\bibliography{references}

\end{document}